%% file: main.tex
\documentclass[final]{cvpr}

\usepackage{times}
\usepackage{epsfig}
\usepackage{graphicx}
\usepackage{amsmath}
\usepackage{amssymb}
\usepackage{caption}

\input{macro}

\def\cvprPaperID{2757} 
\def\confYear{CVPR 2021}

\begin{document}

\title{Fashionpedia-Taste: A Dataset towards Explaining Human Fashion Taste}

\author{
  Mengyun Shi$^{1}$\hspace{8pt}
  Serge Belongie$^{2}$\hspace{8pt}
  Claire Cardie$^{1}$\hspace{8pt}\\
$^{1}$Cornell University
\qquad $^{2}$University of Copenhagen
}

\twocolumn[{
\renewcommand\twocolumn[1][]{#1}
\maketitle
\vspace*{-0.55cm}
\includegraphics[width=\linewidth]{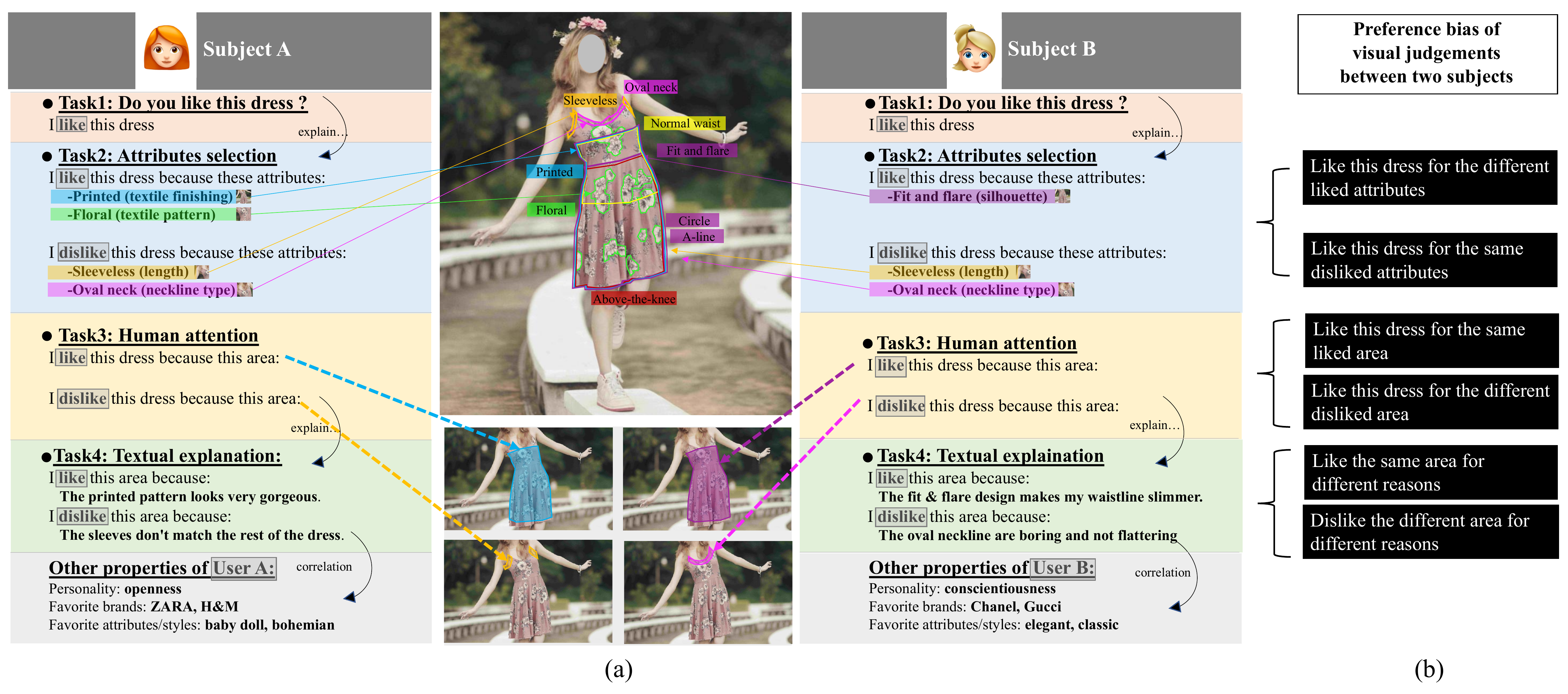}
\captionof{figure}{(a) Explainability \& Reasoning: our Fashionpedia-taste dataset challenges computer vision systems to not only predict whether a subject like a fashion image (task 1), but also provide the explaination from the following 3 perspectives: localized attributes (task 2), human attention (task 3), and caption (task 4); (b) Visual preference bias: even two subjects like a same dress, they can like this dress for totally different reasons (task 2). Similarly, even two subjects like same area of this dress, they can like this area for different reasons (task 3\&4).}
\label{fig:teaser}
\vspace*{0.5cm}
}]

\maketitle


\input{main/sec0_abstract.tex}

\input{main/sec1_intro.tex}

\input{main/sec2_related.tex}

\input{main/sec3_Fashion_Taste_Annotation_from_Subjects.tex}

\input{main/sec4_analysis_1.tex}




\input{main/sec7_conclusion.tex}
\input{main/sec8_acknowledge.tex}


{\small
\bibliographystyle{ieee_fullname}
\bibliography{egbib}
}

\end{document}

%% file: macro.tex
\usepackage{subfigure} 
\usepackage{color}
\usepackage[table]{xcolor}
\usepackage{makecell}
\usepackage{numprint}
\usepackage{multirow}
\usepackage{bbm}
\usepackage{float}
\usepackage{comment}

\newcommand{\ignore}[1]{}

\definecolor{green_im}{rgb}{0.0, 0.5, 0.0}

\newcommand{\cvpara}[1]{\vspace{0.05in}\noindent\textbf{#1}}

\definecolor{citecolor}{RGB}{0, 102, 255}

\usepackage[pagebackref=true,breaklinks=true,colorlinks,citecolor=citecolor,bookmarks=false]{hyperref}

%% file: main/sec0_abstract.tex
\begin{abstract}
    Existing fashion datasets don’t consider the multi-facts that cause a consumer to like or dislike a fashion image.
    Even two consumers like a same fashion image, they could like this image for total different reasons. In this paper, we study the reason why a consumer like a certain fashion image. Towards this goal, we introduce an interpretability dataset, Fashionpedia-taste, consist of rich annotation to explain why a subject like or dislike a fashion image from the following 3 perspectives: 1) localized attributes; 2) human attention; 3) caption. Furthermore, subjects are asked to provide their personal attributes and preference on fashion, such as personality and preferred fashion brands. Our dataset makes it possible for researchers to build computational models to fully understand and interpret consumers’ fashion taste from different humanistic perspectives and modalities. Fashionpedia-taste is available at: \footnote{Fashionpedia project page: \href{https://fashionpedia.github.io/home/}{\texttt{fashionpedia.github.io/home/}}}
   
\end{abstract}

%% file: main/sec1_intro.tex
\section{Introduction}
\label{sec:intro}


Why users click ‘like’ on a fashion image shown on social platforms? They use ‘like’ to express their preference for fashion.
In the context of fashion, apparel represents second highest e-commerce shopping category.
Therefore, fully understand users’ fashion taste could play an important role on the fashion E-commerce.

To increase the chance for a consumer to buy fashion products, a various of recommendation systems have been developed and they have delivered decent results.
However, even a recommendation system makes a correct recommendation for a consumer, does that mean the model really understand the reason why this user like this image? The answer is not clear to us. Furthermore, even two consumers like a same image, they could like this image for totally different reasons, as illustrated in Fig.~\ref{fig:teaser}. To our knowledge, no previous study has explored this problem before.

In this paper, we introduce an explainable fashion taste dataset, Fashionpedia-Taste, which asked the subjects to provide rationale explainations for the reasons that they like or dislike a fashion image from 3 perspectives: 1) localized attribute; 2) human attention; 3) caption.
Additionally, we also collect the extra personal preference information from the subjects, such as preferred dress length, personality, brands, and fine-grained categories while buying a dress. Because these information might also correlate a user’s preference for a fashion image.

The aim of this work is to enable future studies and encourage more investigation to interpretability research of fashion taste and narrow the gap between human and machine understanding of images. The contributions of this work are:
1) an explainable fashion taste dataset consists of 10,000 expressions (like or dislike of an image), 52,962 selected attributes, 20,000 human attentions, and 20,000 captions to explain subjects' fashion taste, over 1500 unique images, and 100 unique subjects; 2) we formalize a new task that not only requires models to predict whether a subject like or dislike a fashion image, but also explain the reasons from 3 different perspectives (localized attribute, human attention, and caption).

%% file: main/sec2_related.tex
\section{Related Work}
\label{sec:related}


\begin{table}
\small
\begin{center}
\resizebox{0.95\columnwidth}{!}{%
\begin{tabular}{ l l l l}
\Xhline{1.0pt}\noalign{\smallskip}
\textbf{Task}
& \textbf{Dataset name}   \\ 
\Xhline{1.0pt}\noalign{\smallskip}
Recognition & iMat~\cite{guo2019imaterialist}, Deep~\cite{liu2016deepfashion}, Clothing~\cite{yamaguchi_parsing_2012}, F-MNIST~\cite{xiao_fashion-mnist:_2017}, \\
 & F-550k~\cite{inoue_multi-label_2017}, F-128~\cite{simo-serra_fashion_2016}, F-14~\cite{takagi_what_2017}, Hipster~\cite{fleet_hipster_2014} \\
\hline
Detection & ModaNet~\cite{zheng_modanet:_2018}, Deep2~\cite{ge_deepfashion2:_2019}, Main~\cite{yu_multi-modal_2017} \\
\hline
Data Mining & Vintage~\cite{hsiao2021culture}, Chic~\cite{yamaguchi_chic_2014}, Ups~\cite{he_ups_2016}, Latent~\cite{hsiao_learning_2017}, Geo~\cite{mall2019geostyle} \\
 & F-144k~\cite{simo-serra_neuroaesthetics_2015}, F-200K~\cite{han_automatic_2017}, Street~\cite{matzen_streetstyle:_2017}, Runway~\cite{vittayakorn_runway_2015} \\
\hline
Retrieval & DARN~\cite{huang_cross-domain_2015}, WTBI~\cite{kiapour_where_2015}, Zappos~\cite{yu2017semantic}, Deep~\cite{liu2016deepfashion} \\
 & Capsule~\cite{hsiao2018creating}, POG~\cite{chen2019pog}, VIBE~\cite{hsiao2020vibe}, IQ~\cite{wu2021fashion}
 \\
\hline 
Attribute & Fashionpedia~\cite{jia2020fashionpedia} \\
Localization & \\
\Xhline{1.0pt}\noalign{\smallskip}
\textbf{Explainability} & \textbf{Fashionpedia-Taste} \\
\textbf{\& Reasoning} & \\
\Xhline{1.0pt}\noalign{\smallskip}
\end{tabular}
}
\caption{Compared to the previous work, Fashionpedia-Taste is the only study that investigates human fashion taste from different humanistic perspectives and modalities.}
\label{tab:literature_review_f_taste}
\vspace{-0.5cm}
\end{center}
\end{table}

\cvpara{Fashion dataset}
Most of the previous fashion datasets focus on recognition, detection, data mining, or retrieval tasks (Table~\ref{tab:literature_review_f_taste}). In the domain of interaction between users and fashion images, Fashion IQ~\cite{wu2021fashion} provides human-generated captions that distinguish similar pair of garment images through natural language feedback. ViBE~\cite{hsiao2020vibe} introduces a dataset to understand users' fashion preference based on her specific body shape. Fashionpedia-Ads studied the correlation between ads and fashion taste among users. Unlike Fashion IQ, ViBE and Fashionpedia-Ads, our dataset focuses on explaining and reasoning on users’ fashion preference based on both visual and textual signal.
Beyond fashion domain, the most relevant work is VCR~\cite{zellers2019recognition}, which requires models to answer correctly and then provide a rationale justificatoin to its answer.
Unlike VCR, our dataset requires models to complete more complicated multi-stage reasoning through different modalities (task1/2/3/4) for subjects’ fashion taste, as illustrated in Fig.~\ref{fig:teaser}.



%% file: main/sec3_Fashion_Taste_Annotation_from_Subjects.tex
\section{Fashion Taste Annotation from Subjects}
\label{sec:sec3_Fashion_Taste_Annotation_from_Subjects}

\cvpara{Subjects and Annotation pipeline}
We recruited 100 female subjects from a U.S. university. Our user annotation process consists of 2 parts: 1) collect subjects' basic information (Sec.~\ref{subsec:userinfo}); 2) collect subjects' fashion taste for given dress length categories (Sec.~\ref{subsec:usertaste}). All the subjects are required to complete these two process.

\subsection{Basic Information Survey}
\label{subsec:userinfo}
This survey is used to collect the subjects': 1) basic information (gender, ethnicity, age); 2) personality; 3) basic fashion preference (favorite fashion brands, fashion attributes and categories); 4) favorite dress length, which is used to determine the images from which dress length should be assigned to each subject for the survey mentioned in Sec.~\ref{subsec:usertaste}.



\cvpara{Personality}
Similar to~\cite{murrugarra2019cross}, we use the 10-item multiple-choice questions to measure subjects' personality.  
We collect personality data because we want to see whether there is a correlation between subjects' personality and their fashion taste.

\cvpara{Basic fashion preference}
We collect users' fashion preference (fashion categories, fashion attributes, and brands) because we are curious whether this self-reported fashion preference is aligned with their preference measured in Sec.~\ref{subsec:usertaste}.

\subsection{Fashion Taste Survey}
\label{subsec:usertaste}

\cvpara{Task Design}
In the fashion taste survey, the subjects are given 100 dress images based on their favourite dress lengths reported in their basic information survey (Sec.~\ref{subsec:userinfo}). They are required to tell whether they like these dresses and provide the reasons why they like or dislike these dresses. For each given dress image, they need to complete the following 4 tasks: 

\begin{itemize}
    \item Task 1: judge whether they like or dislike a given dress.
    \item Task 2-Attribute selection: explain which aspects make them like and dislike a given dress.
    \item Task 3-Human attention: indicate (draw polygons) the regions of the dress that make them like and dislike a given dress. 
    \item Task 4-Textual explanation: explain why the regions they draw from task 3 make them like and dislike a given dress. 
\end{itemize}

\cvpara{Why we design these 4 tasks} Task 2, 3 and 4 allow the subjects to explain their fashion taste from 3 different perspectives and modalities. Task 2 allows the subjects to explain their fashion taste on the perspective of fine-grained attributes. However, Task 2 might miss to capture some information that can only be explained visually. Task 3 is used to address this issue and simulates human gaze capture, allowing the subjects to explain their fashion taste visually. Furthermore, to fully understand the area that subjects draw in task 3, we use Task 4 to allow the subjects to further explain why their draw the areas in Task 3 textually.

Imbalanced likes  and dislikes: we expect it will have big data imbalance if we only ask the subjects to explain the reasons that make them like a dress. To address this issue, we asked users to explain both the aspects that make them like and dislike a given image for task 2, 3, and 4.

%% file: main/sec4_analysis_1.tex
\section{Dataset Analysis}
\label{sec:data_analysis}

\subsection{User annotation analysis}

\cvpara{User basic info}

\begin{figure*}[t]
\centering
\includegraphics[width=\textwidth]{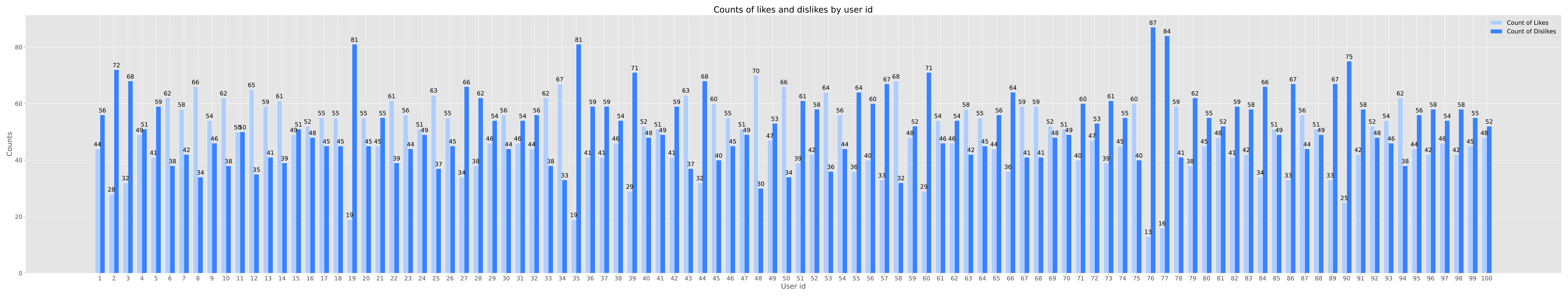}
\caption{Task 1-Count of likes and dislikes by the user. } 
\label{fig:Count of likes and dislikes by the user}
\end{figure*}

\cvpara{Task1-Like / dislike}
We collect 4766 likes and 5234 dislikes, over 1500 unique images, and 100 unique users. The frequency of likes and dislikes selected by each user is shown in Fig.~\ref{fig:Count of likes and dislikes by the user}, indicating most of the users have fairly balanced like and dislike ratio. Balanced like and dislike ratio could potentially help train less biased models.

\cvpara{Task2-Attribute selection}
Table~\ref{tab:Likes and dislikes annotated by the users} breaks down the frequency of liked and disliked attributes selected by each user into different dress lengths. The average liked attributes (3.9397) is nearly 3 times more than disliked attributes (1.4565). This indicates most of the users tend to select the attributes that they like rather than dislike while explaining why they like/dislike a dress.

Table~\ref{tab:The number of total attributes annotated by the users for each fine-grained attribute category} displays the details of annotated attributes into their corresponding fine-grained categories (super-categories). 'Silhouette' contains highest percentage of liked attributes (21.9 \%). This indicates most of users' fashion preference is determined by the shape of a dress. In contrast, 'textile finishing and manufacturing technique' (Tex fini, manu-tech) contains highest percentage of disliked attributes (23.9 \%). This suggests a user could dislike a dress because of this category even she likes the 'silhouette' of a dress.

\begin{figure*}[t]
\centering
\includegraphics[width=\textwidth]
{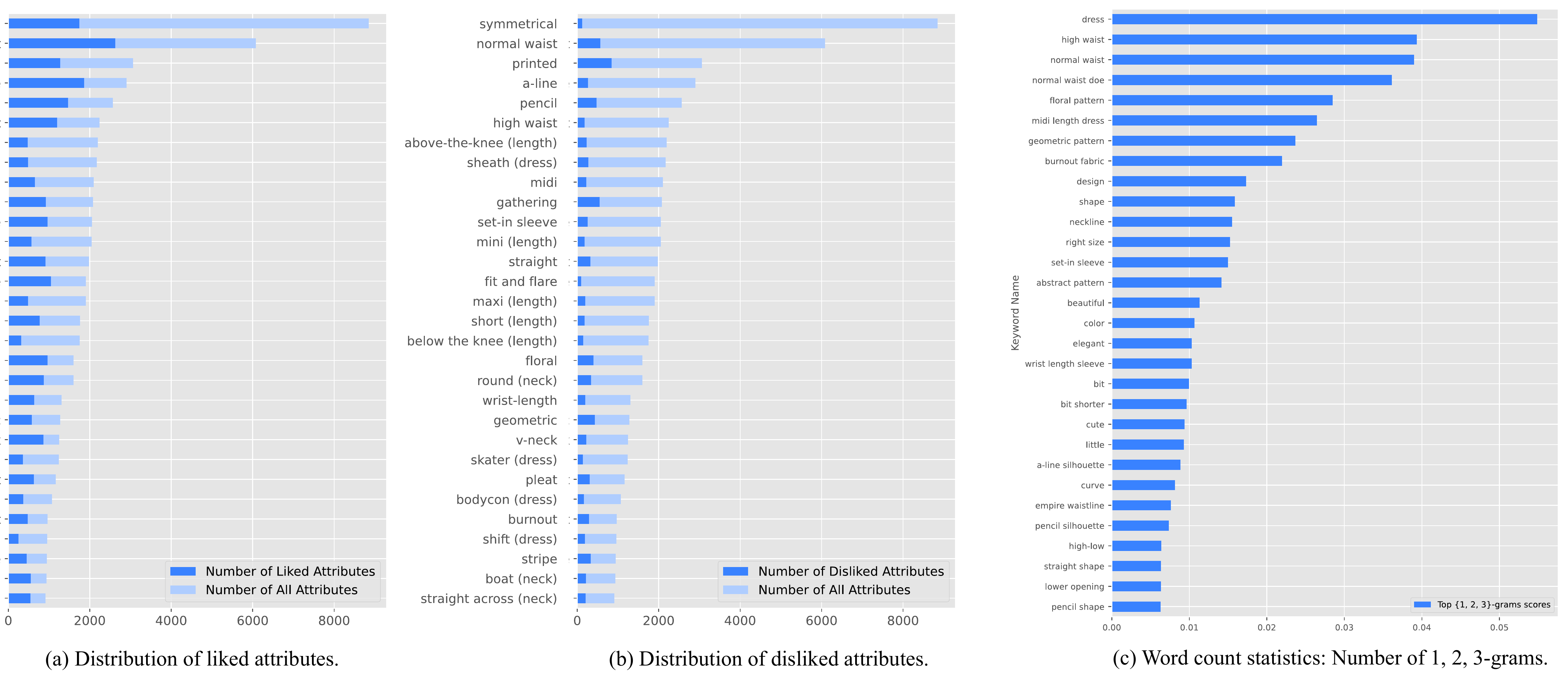}
\caption{Task 2-Distribution of liked and disliked attributes annotated by the subjects \& Task4-Word count statistics} 
\label{fig:Distribution_liked_disliked_attributeTop123-grams_scores}
\end{figure*}

Fig.~\ref{fig:Distribution_liked_disliked_attributeTop123-grams_scores}. (a) \& (b) shows the distribution of liked and disliked attributes annotated by the users. The results show 'printed' and 'normal waist' are the main reason that causes some users to like a dress. However, these attributes can also be the factor that causes users dislike a dress. Because a user could like a certain type of printed pattern but dislike another type of printed pattern. 
To better explain a user's fashion taste, it requires to train a model to not only understand users' preference on a specific attribute, but also the pixel-level pattern on the area that this attribute is located. For this purpose, we asked the users to conduct task 3.

\begin{table}[t]
\small
\begin{center}
\begin{tabular}{ l l l l}
\Xhline{1.0pt}\noalign{\smallskip}
\textbf{Length Type}
&\textbf{Selection type}
&\textbf{\# Total} & \textbf{\# Average}   \\ 
\Xhline{1.0pt}\noalign{\smallskip}
all lengths &Liked &38397 &3.9397 \\
all lengths &Disliked &14565 &1.4565 \\
mini &Liked &7991 &3.89 \\
mini &Disliked &3117 &1.52 \\
above &Liked &8001 &3.63 \\
above &Disliked &3076 &1.39 \\
below &Liked &6312 &3.60 \\
below &Disliked &2561 &1.46 \\
midi &Liked &8463 &4.03 \\
midi &Disliked &3076 &1.46 \\
maxi &Liked &7630 &4.01 \\
maxi &Disliked &2735 &1.43 \\
\Xhline{1.0pt}\noalign{\smallskip}
\end{tabular}
\caption{Task 2-The number of liked and disliked attributes annotated by the users.}
\label{tab:Likes and dislikes annotated by the users}
\vspace{-0.5cm}
\end{center}
\end{table}

\begin{table}[t]
\small
\begin{center}
\begin{tabular}{ l l l l l}
\Xhline{1.0pt}\noalign{\smallskip}
\textbf{Super-category}
&\textbf{\# Liked}
&\textbf{Freq.}
&\textbf{\# Disliked} 
& \textbf{Freq.}  \\
& \textbf{attribute} 
&
&\textbf{attribute} 
&\\
\Xhline{1.0pt}\noalign{\smallskip}
Dress Style & 2385 & 6.2 \% &1132 &7.7 \% \\
Silhouette & 8419 & 21.9 \% &1786 &12.2  \% \\
Textile Pattern & 3194 & 8.3 \% &2031 &13.9  \% \\
Tex fini, manu-tech & 5656 & 14.7 \% &3493 &23.9  \% \\
None-Textile Type & 49 & 0.1 \% &49 &0.33  \% \\
Neckline Style & 6611 & 17.2 \% &2283 &15.6  \% \\
Collar Style & 387 & 1 \% &101 &0.7  \% \\
Lapel Style & 28 & 0.1 \% &4 &0.02  \% \\
Sleeve Style & 2179 & 5.6 \% &726 &4.9  \% \\
Sleeve length & 1917 & 4.9 \% &592 &4.1  \% \\
Pocket Style & 236 & 0.6 \% &84 &0.5  \% \\
Opening Type & 691 & 1.8 \% &453 &3.1  \% \\
Waistline & 4146 & 10.7 \% &849 &5.8  \% \\
Dress length & 2499 & 6.5 \% &982 &6.7  \% \\
\Xhline{1.0pt}\noalign{\smallskip}
\end{tabular}
\caption{Task 2-The number of total attributes annotated by the users for each fine-grained attribute category.}
\label{tab:The number of total attributes annotated by the users for each fine-grained attribute category}
\vspace{-0.5cm}
\end{center}
\end{table}




\cvpara{Task3-Human attention}
Table~\ref{tab:The number of total human gaze annotated by the users for each dress length} shows the number of total human attention annotated by the users for each dress length. The number of annotated attention is evenly distributed across different dress length.

\begin{table}[t]
\small
\begin{center}
\begin{tabular}{ l l l l}
\Xhline{1.0pt}\noalign{\smallskip}
\textbf{Length Type}
& \textbf{\# Total gaze}   \\ 
\Xhline{1.0pt}\noalign{\smallskip}
all lengths & 20000 \\
mini & 4100 \\
above & 4400 \\
below & 3500 \\
midi & 4200 \\
maxi & 3800 \\
\Xhline{1.0pt}\noalign{\smallskip}
\end{tabular}
\caption{Task 3-The number of total human attention annotated by the users for each dress length.}
\label{tab:The number of total human gaze annotated by the users for each dress length}
\vspace{-0.5cm}
\end{center}
\end{table}

\cvpara{Task4-Textual explanation for task 3} Task 4 contains 11.3 words in average per user and 5.6 words in average per subtask (task 4.1 for liked explanation and task 4.2 for disliked explanation).   

Word count statistics: we use SGRank from Textacy~\cite{textacy2002} to calculate the frequency of words. Fig.~\ref{fig:Distribution_liked_disliked_attributeTop123-grams_scores}.(c) shows the most frequent 1, 2, 3 grams for our dataset. 
Waistline (high, normal, and empire waist) and pattern (floral, geometric, abstract) related words are high frequency words used by the users to explain the attention that they draw for Task3. 

Linguistic statistics: we use part-of-speech (POS) tagging from Spacy~\cite{spacy2002} to tag noun, propn, and adj of the captions annotated in Task 4.
Table~\ref{tab:Linguistic statistics: Number of unique words by POS} shows the number of most frequent unique words by POS. We find the most frequent common nouns are more associated with high level description of a dress, such as neckline, pattern and waistline. In contrast, the most frequent proper nouns are more related to detailed description of a dress, such as applique, bead, and peter pan collar. This shows the linguistic diversity of our dataset.


\begin{table}[t]
\small
\begin{center}
\begin{tabular}{ l l l l}
\Xhline{1.0pt}\noalign{\smallskip}
\textbf{POS Type}
& \textbf{Word}   \\ 
\Xhline{1.0pt}\noalign{\smallskip}
Noun & dress, neck, length, neckline, design, shape, pattern, \\
& waist, color, sleeve, curve, waistline, fabric, skirt \\
Propn & maxi, applique, bead, kimono, peter pan, pleat, tiered \\
& halter, dolman, slit, tent, stripe, cutout, cheetah  \\
Adj & elegant, beautiful, nice, cute, high, straight, perfect \\
& floral, loose, graceful, sexy, fit, attractive, charming \\
\Xhline{1.0pt}\noalign{\smallskip}
\end{tabular}
\caption{Task 4-Linguistic statistics: Number of unique words by POS.}
\label{tab:Linguistic statistics: Number of unique words by POS}
\vspace{-0.5cm}
\end{center}
\end{table}

%% file: main/sec7_conclusion.tex
\section{Conclusion}

In this work, we studied the problem of human taste in fashion product image. we introduce an explainable fashion taste dataset, Fashionpedia-Taste, with a purpose to understand fashion taste from 3 perspectives: 1) localized attribute; 2) human attention; 3) caption. The aim of this work is to enable future studies and encourage more investigation to interpretability research of
fashion taste and narrow the gap between human and machine understanding of images.